\pdfoutput=1

\documentclass[11pt]{article}

\usepackage[dvipsnames]{xcolor}

\usepackage[]{acl}

\usepackage{times}
\usepackage{latexsym}
\usepackage{booktabs}
\usepackage{arydshln}
\usepackage{amsmath}
\usepackage{amsfonts}
\usepackage{xspace}

\usepackage[T1]{fontenc}
\usepackage[utf8]{inputenc}
\usepackage{microtype}
\usepackage{graphicx}
\usepackage{subcaption}
\usepackage{enumitem}

\newcommand{\methodname}[1]{\texttt{\textbf{CLCA}}\xspace}

\definecolor{nice-red}{HTML}{E41A1C}
\definecolor{nice-orange}{HTML}{FF7F50} 
\definecolor{nice-green}{HTML}{40B5AD}
\definecolor{nice-blue}{HTML}{6495ED}
\definecolor{nice-purple}{HTML}{B990C9}
\definecolor{gray-blue}{HTML}{92c5de}
\definecolor{gray-red}{HTML}{edb7bd}

\definecolor{dgray}{HTML}{5D6D7E}
\definecolor{dblue}{HTML}{2980B9}
\definecolor{gray}{HTML}{76818c}

\title{Cultural Learning-Based Culture Adaptation of Language Models} 

\author{Chen Cecilia Liu\textsuperscript{1} \and Anna Korhonen\textsuperscript{2} \and Iryna Gurevych\textsuperscript{1}\\
\textsuperscript{1} Ubiquitous Knowledge Processing Lab, \\ 
Department of Computer Science and Hessian Center for AI (hessian.AI), \\
Technical University of Darmstadt\\
\textsuperscript{2} Language Technology Lab, University of Cambridge\\
{\url{www.ukp.tu-darmstadt.de}}\\
}

\begin{document}
\maketitle
\begin{abstract}

Adapting large language models (LLMs) to diverse cultural values is a challenging task, as existing LLMs often reflect the values of specific groups by default, and potentially causing harm to others. In this paper, we present \methodname{}, a novel framework for enhancing LLM alignment with cultural values based on cultural learning.
The framework leverages simulated social interactions to generate conversations in which LLMs engage in role-playing within culturally adapted social scenarios, capturing implicit cultural norms for model fine-tuning. \methodname{} improves cultural value alignment across various model architectures measured using World Value Survey data, demonstrating the effectiveness of our proposed approach. Our results provide early evidence that understanding intent and social interactions can enhance cultural value adaptation in LLMs, highlighting the promise of training approaches based on cultural learning.\footnote{Code: \href{https://github.com/UKPLab/arxiv2025-clca}{CLCA}}
\end{abstract}

\section{Introduction}

Culture has become an increasingly important topic in natural language processing (NLP), particularly following the wide adoption of Large Language Models (LLMs) \cite{DBLP:conf/acl/HershcovichFLLA22, DBLP:journals/corr/abs-2403-15412, culture_survey}. Despite their success, deploying LLMs in real-world applications requires these models to be culturally competent, and adapt to different values and perspectives. However, current LLMs lack such competency across a diverse range of tasks \cite[inter alia]{analysis/cao-etal-2023-assessing,concept/maps/naacl/liu2023, concept/khanuja2024image}, and aligning primarily with WEIRD (Western, Educated, Industrialized, Rich, and Democratic, \citealt{henrich2010weirdest}) values by default, limiting their global applicability.

\begin{figure}
    \centering
    \includegraphics[width=0.86\linewidth]{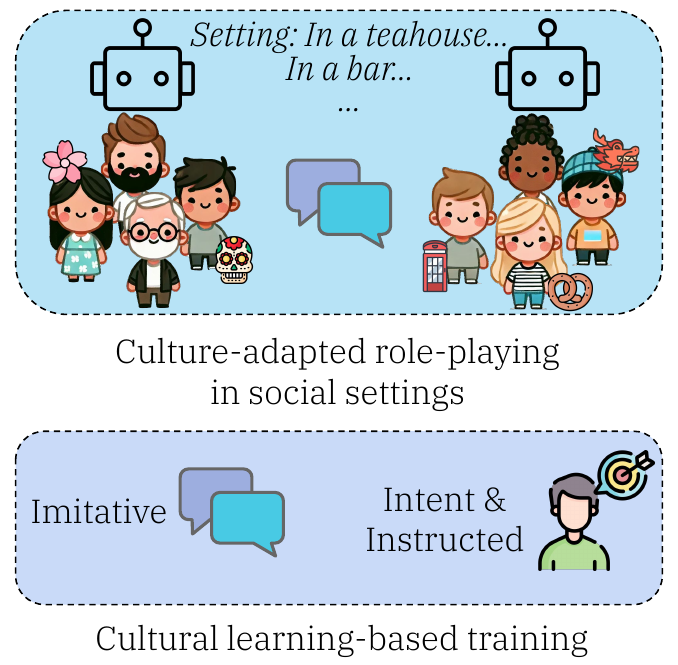}
    \caption{We use culture-adapted role-playing to generate synthetic social interaction conversations. Then, the proposed cultural learning-based framework jointly trains on conversations, intents and their relevance to culture, to improve cultural value alignment.}
    \label{fig:main_small}
\end{figure}

Existing methods for adapting language models to diverse cultural values often rely on prompt engineering \cite{pnas/tao2024cultural, nvm/anthroprompt/acl/abs-2402-13231}. These approaches use demographic information and anthropological reasoning to modify how models respond to human survey questions during inference. However, prompting relies on LLMs already embedding sufficient cultural values during pre-training. \citet{value/cultvalueshiftsft/choenni2024echoes} investigate the impact on diverse cultural value shifts through additional generic pre-training corpora. The study reveals that while training on such data may embed additional cultural signals into models, it often falls short in achieving controlled adaptation to specific cultures. These findings emphasize the need for further research to enhance the cultural value alignment of LLMs. 

Recent studies \cite{Bhoopchand2023LearningFI, cl/natmi/DuenezGuzmanSWML23} show the importance of cultural learning in training intelligent systems. Cultural\footnote{\emph{Culture} here encompasses a broader scope than its typical use in NLP. It includes fundamental human processes that are integral to society and can be transmitted, such as scientific discoveries, hunting practices, language learning, and more.} learning \cite{culturallearning/tomasello1993cultural, culturallearningredux/tomasello2016cultural, tomasello2019becoming, henrich2016secret, heyes2017does} enables humans to acquire knowledge and behaviours through social interactions and observation within a shared cultural context,\footnote{Our cultural values are often reflected in our actions, words, and social behaviours.} facilitating cultural transmission and cultural evolution in humans across generations. 

Key aspects of cultural learning highlight that culture is acquired through mechanisms such as imitation and instruction, along with the ability for intent understanding (or ``mind-reading'', \citealt{premack1978does}), and enables individuals to internalize behaviours and values from their communities through social interactions. While prior research in NLP has explored the sociality and social interactions of LLMs \cite[inter alia]{roleplaying/conf/uist/ParkPCMLB22, sociality/corr/abs-2305-16960, sharma-etal-2024-investigating, emnlp/roleplaydoh/louie-etal-2024-roleplay, cl/socialbench/chen-etal-2024-socialbench, icml/debate/Du00TM24} --- including areas such as decision-making and human-AI collaboration --- there has been limited attention given to leveraging concepts in cultural learning (\S\ref{sec:cl}) for behaviour-driven cultural value adaptation. Inspired by this human-centric view, we propose a \textbf{C}ultural \textbf{L}earning-based framework for \textbf{C}ulture \textbf{A}daptation (\methodname{}, Figure \ref{fig:main_small}\footnote{Icons in Figures are from Flaticon.com or created with the assistance of DALL-E.}), adapting LLMs to different cultural values by leveraging simulated social interactions. 
By incorporating elements of imitative learning, instructed learning, and intent understanding, \methodname{} improves cultural value alignment across multiple LLMs.

\begin{figure*}
    \centering
    \includegraphics[width=0.94\linewidth]{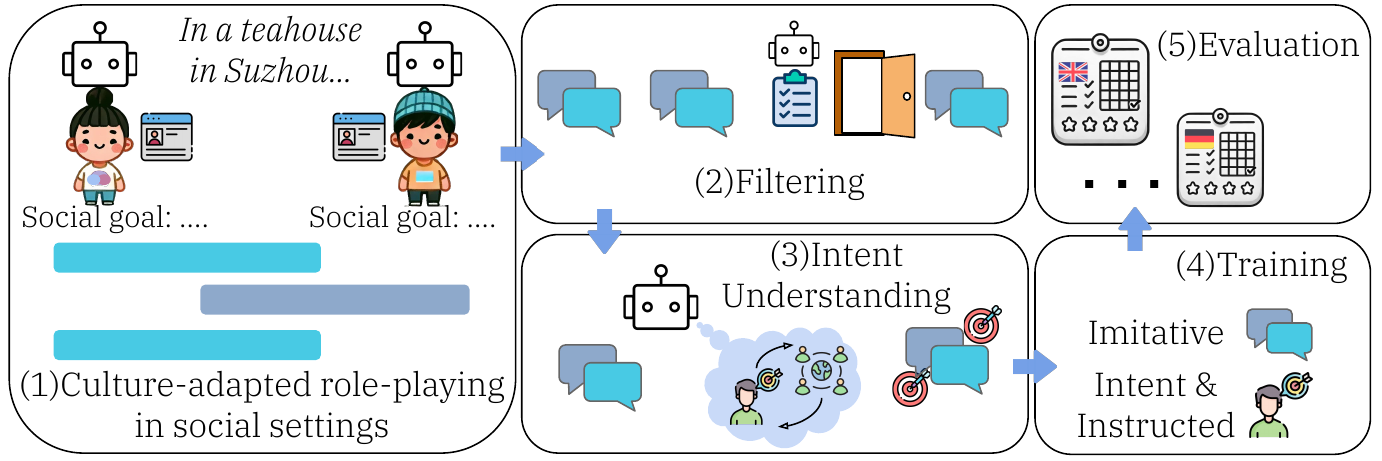}
    \caption{(1) The framework first automatically generates conversations through culture-adapted role-playing in social settings. (2) These conversations are then filtered using GPT models to ensure quality and relevance. (3) The filtered data is labelled with free-text intents. (4) Both the conversation and intent data are integrated into a cultural learning-based training process (\methodname{}). (5) The resulting models are evaluated using the World Values Survey.}
    \label{fig:main}
\end{figure*}

\noindent{\textbf{Contributions.}} To summarize: \textbf{1)} We propose \methodname{} for cultural value adaptation by leveraging synthetic conversations generated through simulation (i.e., role-playing) of LLMs in generated social situations. \textbf{2)} We show that simulated social conversations effectively improve LLMs' response alignment with survey questions across different cultures and various models. \textbf{3)} Through extensive ablation studies, we demonstrate that social interaction data and intent understanding are essential for adapting models through cultural learning.

\section{Related Work}\label{sec:related_work}

\noindent{\textbf{Adapting LLMs to Cultural Values.}} 
Recent studies show the effectiveness of role-playing prompts in improving cultural value alignment in LLMs. For instance, \citet{pnas/tao2024cultural} demonstrate that prompting LLMs to role-play as (generic) individuals from specific cultures effectively improves their cultural value alignment.  
While lightweight, this relies on the assumption that a model has already acquired sufficient cultural values. Similarly, \citet{nvm/anthroprompt/acl/abs-2402-13231} introduce anthropological reasoning prompting with fine-grained demographic information and improved alignment with Arabic cultural values, as assessed using World Values Survey (WVS) data. These findings suggest that role-playing influences the evaluation of cultural values, allowing targeted adaptation of models during evaluations. Alternatively, studies such as \citet{cultureLLM/corr/abs-2402-10946, li2024culturepark} focus on leveraging explicit value data to adapt downstream tasks, either through direct tuning or synthetic data based on value surveys. This approach leads to explicit, value-driven behavioural changes, which differ from ours (i.e., behaviour-driven value changes).

Close to ours, \citet{value/cultvalueshiftsft/choenni2024echoes} examine the impact of fine-tuning with different pre-training corpora \cite{DBLP:journals/lre/Christodoulopoulos15, tacl/flores101/goyal-etal-2022-flores} on cultural value shifts. Their results suggest that the semantic content (e.g., news, Bible) alone of the fine-tuning data does not effectively induce controlled value alignment across various cultures. Our work focuses on utilizing simulated social interactions, inspired by cultural learning theories from evolutionary anthropology and psychology.

\noindent{\textbf{Synthetic Data Generation \& Simulations in Social Settings.}} Generating synthetic data with LLMs is a promising way to enhance various model capabilities \cite{nvm/prosocial/emnlp/kim-etal-2022-prosocialdialog, yue-etal-2023-synthetic, lu-etal-2024-mathgenie}. LLMs can effectively role-play characters \cite{demographic/Argyle_Busby_Fulda_Gubler_Rytting_Wingate_2023, roleplaying/conf/uist/ParkPCMLB22}, for both domain-general and domain-specific applications \cite[inter alia]{icml/debate/Du00TM24, collaboration/acl/zhang-etal-2024-exploring, conflicresolution/chi/ShaikhCGYB24, emnlp/roleplaydoh/louie-etal-2024-roleplay}.
While roleplay-based synthetic data improves LLM performance in social contexts \cite{zhou2024sotopia, acl/sotopiapi/wang-etal-2024-sotopia, DBLP:journals/corr/abs-2407-03974}, prior work does not address adaptation to different cultural values or specifically examine cultural learning.

\section{Cultural Learning}\label{sec:cl}

Cultural learning is a general concept from anthropology and psychology \cite[inter alia]{culturallearning/tomasello1993cultural,  culturallearningredux/tomasello2016cultural,tomasello2019becoming,henrich2003evolution,henrich2016secret,heyes2017does,heyes2018cognitive} that refers to the process by which individuals acquire behaviours, knowledge, and other aspects of ``culture'' from their social environment. It is critical in shaping human social values and enabling the transmission of culture across generations. 

There are three primary forms of cultural learning \cite{culturallearning/tomasello1993cultural}: 1. imitative learning, 2. instructed learning and 3. collaborative learning. This work focuses on imitative and instructed learning, as they represent the foundational forms through which individuals first acquire culture (i.e., transmission of culture).\footnote{We will leave collaborative learning in future work as it involves the co-construction or co-creation of (new) culture \cite{culturallearning/tomasello1993cultural}. This approach is less suited to our current focus, which centers on adapting to existing cultures.} We provide a brief description of each form below.

\noindent{\textbf{Imitative Learning.}} This involves observing and replicating the actions of others (often adults or experts). In robotics and reinforcement learning, it is implemented through methods such as imitation learning \cite{imitation_learning/ftrob/OsaPNBA018}, behaviour cloning \cite{behaviour_cloning/ijcai/TorabiWS18}, or supervised fine-tuning like in NLP. Imitative learning is key to skill acquisition, particularly in childhood, as individuals learn by mimicking behaviours without necessarily understanding the underlying intent.

\noindent{\textbf{Instructed Learning.}} In this form, the cultural knowledge or practices are explicitly conveyed or demonstrated. Instructed learning allows learners to acquire essential cultural practices within a limited timeframe. 

One important factor in cultural learning is the ability to understand the intentions of others during interactions. In imitative learning, understanding intention can help differentiate between actions that are essential to a task and those that are incidental. Similarly, in instructed learning, understanding the intent behind instructions enhances the learner's ability to generalize and apply knowledge in various contexts. 

\section{Method}\label{sec:method}

Our overall adaptation framework is in Figure \ref{fig:main}.

\subsection{Social Data Generation}
\noindent{\textbf{Culture-Adapted Social Scenarios.}} We use the setup of text descriptions of social scenarios, character profiles and corresponding social goals following setup in Sotopia \cite{zhou2024sotopia, acl/sotopiapi/wang-etal-2024-sotopia}. To make them appropriate for culture-based interactions, we perform automatic culture adaptations of social settings in \cite{acl/sotopiapi/wang-etal-2024-sotopia} using a GPT-4 model (prompts in Appendix \ref{app:prompts}), as well as generating new scenarios based on social and cultural norms from Social Chemistry \cite{forbes-etal-2020-social} and Culture Atlas.\footnote{\url{https://culturalatlas.sbs.com.au/}} Each social task contains a setting, two participant profiles (including name, age, gender and occupation), and their respective private social goals for the interaction. After the adaptation, participant names are localized (e.g., from \emph{Anthony} to \emph{Henrik} or \emph{Kenji}) and settings are adapted (e.g., from \emph{Alps} to \emph{Yunnan}, or from \emph{a bar in London} to \emph{a teahouse in Suzhou}).

\noindent{\textbf{Interaction Data Generation.}}
Following \citet{simulation/fantasy/zhou-etal-2024-real, zhou2024sotopia}, two LLMs are role-playing the participants (in ``agent mode''). During the interaction, the shared information is the setting (e.g., ``a mentor and mentee team up discussing a research project'' ), and participants' basic information (e.g., ``Jie Li'', ``45 / female'', ``a senior researcher'').  The social goals and secrets are only visible to each LLM (e.g., ``ensure that the project reflects university's priority and interests''). The data generation process is guided by incorporating cultural context from Hofstede's cultural dimensions \cite{hofstede2010cultures} and Inglehart–Welzel cultural map \cite{inglehart2005christian} into the system prompt (see Appendix~\ref{app:cultural_guide}).

Unlike the prior work \cite{zhou2024sotopia, acl/sotopiapi/wang-etal-2024-sotopia}, the completion rate of these goals in interaction is not relevant to our study. Instead, we focus on the implicit social and cultural values during interactions and use them for cultural value adaptation (an example conversation in Table \ref{tab:example}).

\noindent{\textbf{Filtering.}}
To ensure the data quality, we filter the generated synthetic data by using LLM-as-a-Judge \cite{nips/chatbotarena/ZhengC00WZL0LXZ23,icml/ultrafeedback/CuiY0YH0NXXL0024, iclr/prometheus/KimS0JLLYSKTS24}. We create a two-step rubric-based approach with a model verbalizing its confidence based on prior research \cite[inter alia]{tmlr/verbaluncert/LinHE22, aistats/verbuncert/TanneruAL24, emnlp/uncertainty/dong-etal-2024-llm, iclr/verbuncert/XiongHLLFHH24}. 

We evaluate an entire conversation based on two aspects with confidence: 1. general generation quality, and 2. cultural adherentness. Based on these evaluations, we ask the model to make a meta-evaluation critique on the quality of evaluation and output its confidence (prompts in Appendix \ref{app:prompts}). 

We generate data twice for each social scenario and apply the filtering process. Data labelled with high-confidence bad ``meta-evaluation'' or ``general generation quality'' are discarded. Table \ref{tab:data_stats} presents the resulting data statistics. In this work, we use LLM-as-a-Judge as a proxy for data quality, and we provide a qualitative analysis in Appendix \ref{app:data_quality}.

\noindent{\textbf{Intent Generation.}}
After generating the conversations, the model identifies the free-text intent of each conversational turn based on the history and evaluates its relevance to social and cultural expectations.\footnote{These are for general intentions understanding, distinct from the fixed category intent predictions \cite{intent/chiir/Qu0CZTQ19} or open-world intent discovery \cite{intent/aaai/Zhang0LL21,intent/zhang-etal-2022-new} in dialogue tasks.} Two example intents are in Table \ref{tab:intent} (prompt in Table \ref{tab:intent_prompt} and a detailed example in Table \ref{tab:example}). An intent may be generic (e.g., greeting or signalling the end of the conversation) or reflect culturally specific expectations. When the intent is annotated with culture-specific expectations, we take this form as ``instruction'' (as in instructed learning, introduced in \S\ref{sec:cl}), as it conveys the expected behaviour in a particular culture. 

\begin{table}[ht]
\centering
\resizebox{0.93\columnwidth}{!}{
\begin{tabular}{p{\columnwidth}} 
\toprule
Example Intents \\
\midrule
\textcolor{nice-blue}{\textit{Generic:}} To verify the recipient's identity and return the misdelivered package to its rightful owner.\\
\textcolor{nice-purple}{\textit{Cultural:}} To politely and professionally express interest in Wang Lei's project while maintaining a humble and respectful demeanour, as is expected in Chinese culture when interacting with someone of higher social status or age.\\

\bottomrule
\end{tabular}}
\caption{Generated intent examples.}
\label{tab:intent}
\end{table}

\subsection{Cultural learning-Based Culture Adaptation (\methodname{})}
To enhance the cultural value alignment of LLM, we use a multi-task training approach leveraging the generated data. The training process consists of two tasks: 1. multi-turn conversation, and 2. intent understanding, with respect to cultural and social expectations. 

\noindent{\textbf{Multi-Turn Converstaion Training.}} This task mirrors imitative learning in cultural learning, designed to improve the model's ability to handle contextually rich conversations in social settings. During training, each conversation is used twice (once from each participant's perspective), so the model learns appropriate responses by switching perspectives. 

\noindent{\textbf{Intent Understanding.}} This task focuses on generating the underlying intention of the conversation turn while learning its relevance to social and cultural expectations. This mirrors the instructed learning and intent understanding in cultural learning. During training, the model is provided with contextual information about the social setting and the conversation but does not receive explicit prompts to role-play. This training helps the model handle culturally sensitive scenarios.

By combining these two tasks, our approach is equipped with two basic forms of cultural learning.

\section{Experimental Setup}

\subsection{World Values Survey (WVS) and Evaluation}\label{sec:wvs_eval}
Following the evaluation setup in \citet{nvm/anthroprompt/acl/abs-2402-13231} for measuring cultural values in LLM, we conducted an evaluation using the WVS~\cite{haerpfer2022wvswave7}. The WVS is a survey for public opinions (i.e., cultural values) on a wide range of topics such as economic developments, and religious beliefs across various countries (i.e., geo-political cultures). It is widely used in sociological research to assess cultural shifts and became popular recently in NLP for cultural value evaluations \cite{corr/xlingwvs/abs-2203-13722, nvm/anthroprompt/acl/abs-2402-13231, value/cultvalueshiftsft/choenni2024echoes}. The WVS uses a representative sample of each country's general population across various demographics. It contains questions spanning 13 categories, such as \emph{Social Capital, Trust \& Organizational Membership} or \emph{Security} (Table \ref{tab:wvs_cat} for a complete list). 

In this work, we used the 7th version of the survey (conducted from 2017 - 2020) for five different (geo-political) cultures: United Kindom (UK), China, Germany, Mexico, and Japan. We use all questions from the \emph{Social Values, Norms, Stereotypes} category (44 questions per culture), based on an implementation in WorldValueBench~\cite{nvm/worldvaluesbench/zhao-etal-2024-worldvaluesbench-large}. This category is the most relevant as it closely aligns with our data generation process, which is grounded in social and cultural norms.

To simulate the model's response as a member of a specific cultural group, we utilize the demographic information of survey respondents in WVS, similar to \citet{nvm/anthroprompt/acl/abs-2402-13231}. In this context, we refer to these profiles as \emph{personas} to distinguish them from the character profiles used in our data generation process. These personas are then integrated into the model as the system prompts during evaluation. The information included in the personas is in Table \ref{tab:demo}. 
The questions from the survey are provided to the model as the user prompt, and the template is in Table \ref{tab:user_prompt}. We sample 1000 personas per culture randomly without replacement (a total of 220k questions evaluated per model for all cultures) for evaluation. The survey, originally in English, is further translated for multilingual evaluation (\S\ref{sec:xling}) using the GPT-4 model.

\subsection{Models}
We evaluate the adaptation of the following open source state-of-the-art LLMs: Llama~\cite{llm/llama/abs-2302-13971, llama3/corr/abs-2407-21783} - 3.2 1B/3B, 3.1 8B; Mistral~\cite{mistral} - v0.3 7B; Qwen~\cite{qwen2p5} - 2.5 0.5B/1.5B/7B. Here, the Llama and Qwen models are multilingual. We experiment with all \emph{instruction-tuned} models, due to their performant instruction-following and conversation abilities, as well as their closeness to the realistic usage scenarios (base models are unlikely to be used outside of academic research). 

\subsection{Methods}\label{subsec:method}
\noindent{\textbf{Persona.}} Zero-shot evaluation baseline using the personas described in Table \ref{tab:demo}. There is \emph{no suffix} for this variant in the results tables, and we also refer to this as the \texttt{Standard} evaluation in all figures.

\noindent{\textbf{Cultural.}} Cultural prompting \cite[suffix: \texttt{cultural}]{pnas/tao2024cultural} uses culture-specific prompts but excludes any demographics (i.e., same prompt per culture), serving as another baseline.

We do not compare with existing training-based methods (e.g., \citealt{cultureLLM/corr/abs-2402-10946}) due to differences in goals, as discussed in \S\ref{sec:related_work}. Further, their training data serves as evaluation data in our setting.

\noindent{\methodname{}.} In this work, we aim to enhance the cultural value alignment of smaller models by leveraging the Llama3.1 70B model as the source for conversation generation. Llama3.1 70B is selected for its role-playing capabilities and its suitability for the investigation of cultural learning-based adaptation, where smaller, weaker models learn and adapt by observing ``expert'' behaviour demonstrated by larger models. We use a GPT-4 model \cite{llm/nips/Ouyang0JAWMZASR22} as the judge for data filtering. We use LoRA \cite{DBLP:conf/iclr/HuSWALWWC22} adapters for adaptations (hyperparameters in Appendix \ref{app:hparams}). The evaluation uses the same persona prompts described in \S\ref{sec:wvs_eval}.

\begin{table*}[]
\centering
\resizebox{0.77\linewidth}{!}{
\begin{tabular}{l ccccc l}
\toprule
 & China & Germany & UK & Mexico & Japan & Avg. \texttt{KL-D} \textcolor{SeaGreen}{$\downarrow$} \\
\midrule
Llama3.1 8B &  0.5958 & 0.6717 & 0.6268 & 0.5391 & 0.5721 & 0.6011 \\
Llama3.1 8B$_{\texttt{cultural}}$ & 0.5881 & 0.6690 & 0.6431 & 0.5437 & 0.5660 & 0.6020 \\
Llama3.1 8B$_{\methodname{}}$ & 0.5462 & 0.4935 & 0.5510 & 0.4630 & 0.5024
 & \textbf{0.5112} \textcolor{SeaGreen}{$_{\Delta0.0899}$} \\
\hdashline
Llama3.2 3B & 0.6174 & 0.6903 & 0.6631 & 0.5667 & 0.6221 & 0.6319  \\
Llama3.2 3B$_{\texttt{cultural}}$ & 0.5996 & 0.6729 & 0.6375 & 0.5569 & 0.6042 & 0.6142 \\
Llama3.2 3B$_{\methodname{}}$ & 0.5337 & 0.6732 & 0.6695 & 0.5525 & 0.6100 & \textbf{0.6078} \textcolor{SeaGreen}{$_{\Delta0.0241}$}\\
\hdashline
Llama3.2 1B & 0.5936 & 0.6479 & 0.6384 & 0.5584 & 0.6024 & 0.6081 \\
Llama3.2 1B$_{\texttt{cultural}}$ &  0.5905 & 0.6840 & 0.6675 & 0.5209 & 0.6664 & 0.6259 \\
Llama3.2 1B$_{\methodname{}}$  & 0.5671 & 0.6208 & 0.6348 & 0.5683 & 0.5743 & \textbf{0.5931} \textcolor{SeaGreen}{$_{\Delta0.0150}$}\\ 

\midrule
Qwen2.5 7B & 0.5692 & 0.4610 & 0.4221 & 0.4509 & 0.5053 & \textbf{0.4817} \\
Qwen2.5 7B$_{\texttt{cultural}}$ & 0.5984 & 0.5051 & 0.5355 & 0.4961 & 0.5467 & 0.5364 \\
Qwen2.5 7B$_{\methodname{}}$ & 0.5917 & 0.4605 & 0.4439 & 0.4390 & 0.5047 & 0.4880 \textcolor{orange}{$_{-\Delta0.0063}$}\\

\hdashline
Qwen2.5 1.5B &   0.6315 & 0.6069 & 0.6040 & 0.5134 & 0.6225 & 0.5956\\
Qwen2.5 1.5B$_{\texttt{cultural}}$ & 0.6271 & 0.6406 & 0.6540 & 0.5476 & 0.6343 & 0.6207 \\
Qwen2.5 1.5B$_{\methodname{}}$ & 0.5614 & 0.4895 & 0.6414 & 0.4559 & 0.6129 & \textbf{0.5522} \textcolor{SeaGreen}{$_{\Delta0.0434}$}\\

\hdashline
Qwen2.5 0.5B &   0.6381 & 0.5589 & 0.5205 & 0.5192 & 0.6373 & 0.5748\\
Qwen2.5 0.5B$_{\texttt{cultural}}$ & 0.5661 & 0.6382 & 0.6093 & 0.5305 & 0.5818 & 0.5852  \\
Qwen2.5 0.5B$_{\methodname{}}$ & 0.6130 & 0.5173 & 0.5061 & 0.4428 & 0.5794 & \textbf{0.5317} \textcolor{SeaGreen}{$_{\Delta0.0431}$}\\

\midrule
Mistral-v0.3 7B &  0.6216 & 0.6414 & 0.6249 & 0.5069 & 0.6458 & 0.6081 \\
Mistral-v0.3 7B$_{\texttt{cultural}}$ & 0.6155 & 0.6733 & 0.6553 & 0.5219 & 0.6475 & 0.6227  \\
Mistral-v0.3 7B$_{\methodname{}}$ & 0.6171 & 0.6407 & 0.6178 & 0.5074 & 0.6341 & \textbf{0.6034} \textcolor{SeaGreen}{$_{\Delta0.0047}$}\\

\bottomrule
\end{tabular}}
\caption{The Kullback–Leibler Divergence (KL-D) between the distribution of predicted answers and the distribution of the ground truth answers from the WVS survey of various models on different cultures. All models are instruction-tuned, the green arrow indicates the lower the KL-D the better, and the best average result is in bold. Delta is calculated with respect to the persona baseline (no suffix in the table) since they use the same evaluation prompts.}
\label{tab:main}
\end{table*}

\subsection{Metrics} 
We measure cultural value alignment using two metrics: one at the culture level and one at the individual level (i.e., simulated persona level). While the primary goal of our work is to achieve adaptation at the culture level (i.e., over distributions of answers for a culture), it is also crucial to assess individual-level alignment to avoid issues like improving culture-level alignment while individuals hold swapped answers.

\noindent{\textbf{Kullback–Leibler Divergence.}} To evaluate the similarity between the predicted answer distributions and the ground truth from the survey, we report the culture-level Kullback–Leibler Divergence (KL-D) \footnote{Alternativly, a symmetric metric can also be used, such as Jensen-Shannon Distance (Appendix \ref{app:jsd}).} as follows:

\[
D_{\text{KL}}(P; Q) = \frac{1}{M}\sum_{i=1}^M \sum_{k=1}^{K(i)} P_i(k) \log \frac{P_i(k)}{Q_i(k)},
\]
\noindent
where \(P_i(k)\) represents the probability of the $k$-th answer for question \(i\), and \(Q_i(k)\) represents the probability of the ground truth (i.e., from survey) for the same question and answer. $K(i)$ is the number of answers for question $i$. $M$ is the number of questions used for evaluation (same per culture). We add a category for safeguarded answers when calculating the KL-D, which is a more stringent measure (i.e., assuming all the safeguarded answers are wrong). The best possible KL-D is 0 when two distributions are identical.

\noindent{\textbf{Individual-level Accuracy.}} It is defined as:

\[
\text{Accuracy} = \frac{1}{N} \sum_{n=1}^{N} \left( \frac{1}{M} \sum_{i=1}^{M} \mathbb{I}(\hat{y}_n^i, y_n^i) \right),
\]

where
\[
\mathbb{I}(\hat{y}_n^i, y_n^i) =
\begin{cases}
1 & \text{if } \hat{y}_n^i = y_n^i, \\
0 & \text{otherwise}.
\end{cases}
\]
$\hat{y}_n^i$ is the model predicted answer, $N$ is the total number of personas. The best possible value is 1.

\section{Results and Discussion}\label{sec:results}

\begin{figure*}
    \centering
    \includegraphics[width=0.92\linewidth]{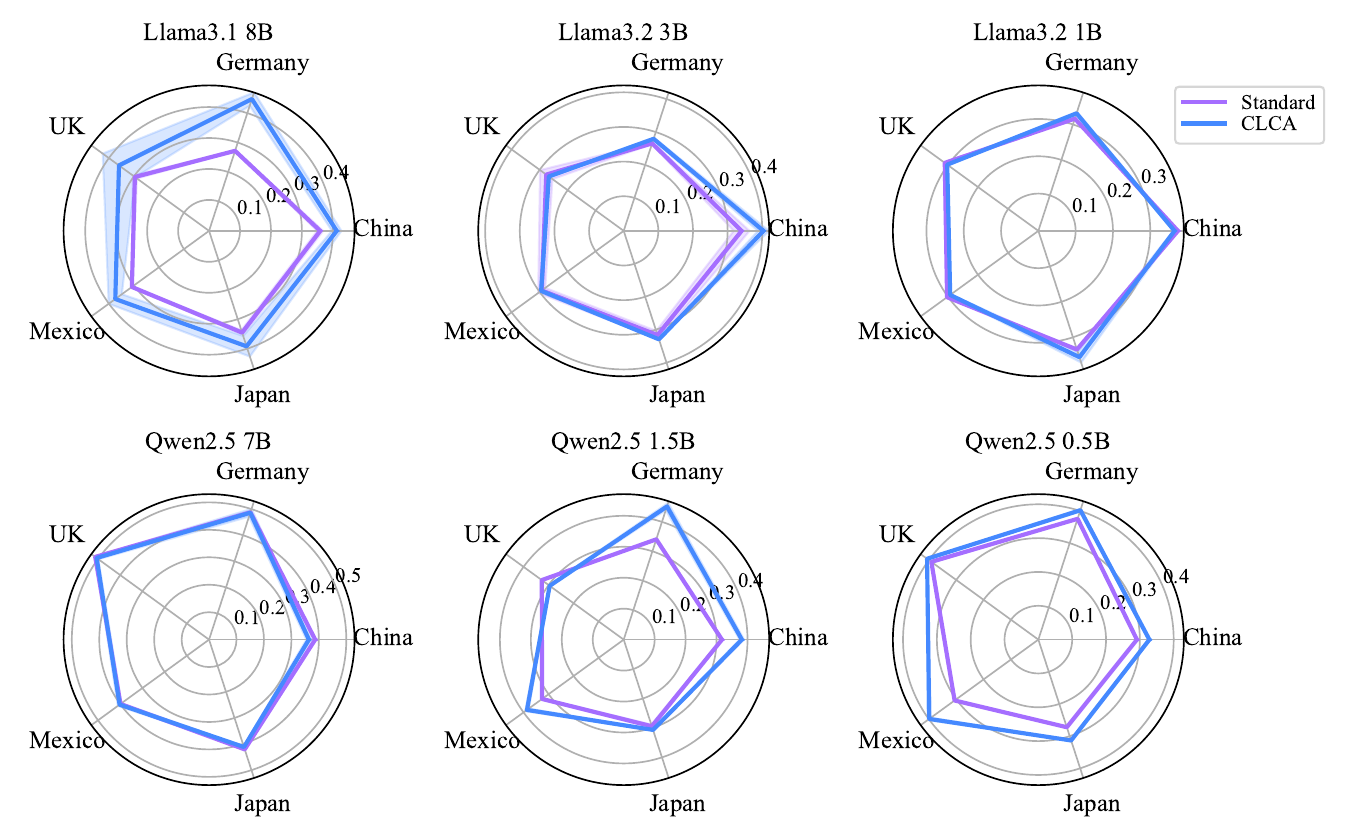}
    \vspace{-1em}
    \caption{The individual-level accuracy (the higher the better) of \methodname{} versus zero-shot results of the persona baseline (\texttt{Standard}, described in \S\ref{subsec:method}) against the ground truth answers from the survey for different cultures. Mistral results in Figure \ref{fig:mistral}, and averages for all models in Table \ref{tab:acc_main} in the Appendix. All models are instruction-tuned.}
    \label{fig:radar}
\end{figure*}

\subsection{Cultural Learning Aligns Models to Surveys}

Table \ref{tab:main} shows the KL-D across different cultures and models. In general, the persona baseline (no suffix) tends to perform better than the cultural baseline. 
Our method, \methodname{}, consistently outperforms the persona baseline across various model sizes and types, with the exception of Qwen2.5 7B. Notably, the largest improvement is over Llama3.1 8B with a reduction of 0.0899 in KL-D. Further, we do not observe clear scaling trends in Qwen models. However, larger Llama models appear to be more adaptable.

While our goal is to improve culture-level alignment, it is important to verify if individual-level accuracy improves. Figure \ref{fig:radar} shows the results across different models and cultures for the persona baseline (i.e., \texttt{Standard}) and \methodname{}. Similarly, the largest improvement is observed for the Llama3.1 8B model across all cultures. 

\subsection{Social Interaction Plays a Significant Role}
A key question is whether social interaction data is important for the controlled improvement of culture alignment. To validate this, we perform two experiments with mathematical reasoning datasets that exhibit \emph{minimal} cultural and social conventions in a typical social interaction setting. The first experiment utilizes the GSM8K dataset \cite{cobbe2021gsm8k}, which consists of single-question mathematical reasoning problems with corresponding answers. We reformulate this as a one-turn conversation where a user poses a question, and the model provides the answer (left panel in Figure \ref{fig:gsm8k}). The second experiment employs the MathChat dataset \cite{mathchat}, a multi-turn conversational dataset for mathematical reasoning. It begins with a single question and answer, followed by additional follow-up questions about the problem (right panel in Figure \ref{fig:gsm8k}). This multi-turn nature mirrors our synthetically generated conversations. We train the Llama3.1 8B using the same format, system prompt, and personas as in previous experiments, but replace the simulated conversations with mathematical reasoning datasets.

\begin{table}[]
    \centering
    \resizebox{0.88\linewidth}{!}{
    \begin{tabular}{l cc}
\toprule
Model & Acc \textcolor{SeaGreen}{$\uparrow$} & KL-D \textcolor{SeaGreen}{$\downarrow$}\\
\midrule
    Llama3.1 8B & 0.3162 &  0.6011\\
    Llama3.1 8B$_{\methodname{}}$ & \textbf{0.3973} & \textbf{0.5112}\\
    \hdashline
    Llama3.1 8B$_\texttt{GSM8K}$  & 0.3287 & 0.5902\\
    Llama3.1 8B$_\texttt{MathChat}$  & 0.3260 & 0.5818\\
    \bottomrule
    \end{tabular}
    }
    \caption{Comparison of Llama3.1 8B model trained with reasoning-only datasets versus training with social conversations. All models are instruction-tuned, the direction of the arrows indicates if the values should be maximized or minimized.}
    \label{tab:ablation_gsm8k}
\end{table}

Table \ref{tab:ablation_gsm8k} shows that training exclusively on mathematical reasoning datasets improves the results by a small margin. This is expected, as any update in model weights affects the model's predictions. However, compared to social interaction data, this adjustment has a minimal effect on aligning the model's evaluation with WVS data. 
We conducted two additional experiments using cultural knowledge data presented in a conversational format (Appendix \ref{app:ablation}, Table \ref{tab:ablation_gsm8k2}) to better isolate the effect of social interactions. These experiments confirmed our original conclusion.

\begin{figure*}[h!]
    \centering
    \begin{subfigure}{0.49\linewidth}
        \includegraphics[width=\textwidth]{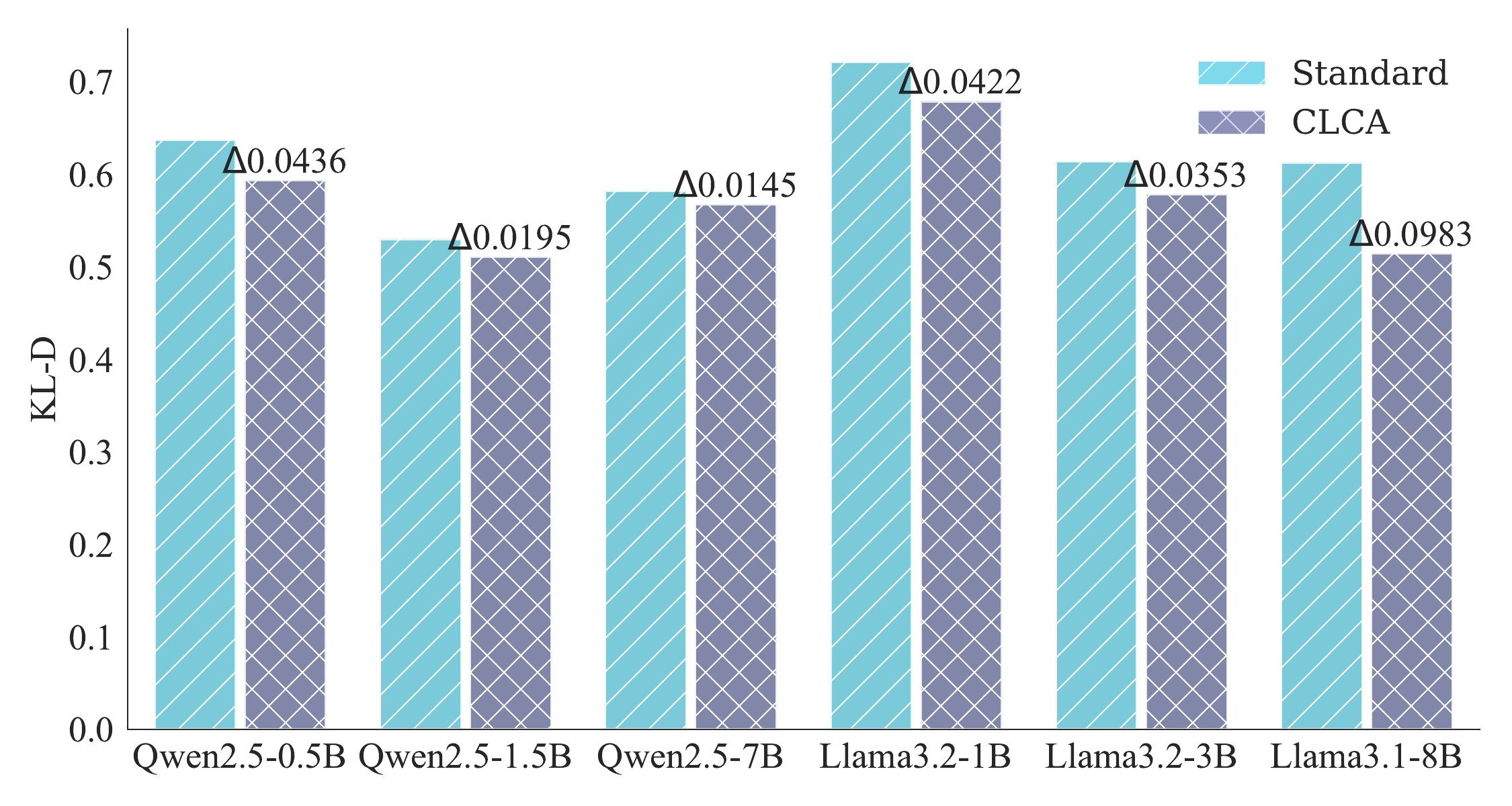}
        \caption{Kullback–Leibler Divergence (KL-D, lower is better) between the model prediction and WVS data.}
        \label{fig:llama_xling_kld}
    \end{subfigure}
    \hfill
     \begin{subfigure}{0.49\linewidth}
        \includegraphics[width=\textwidth]{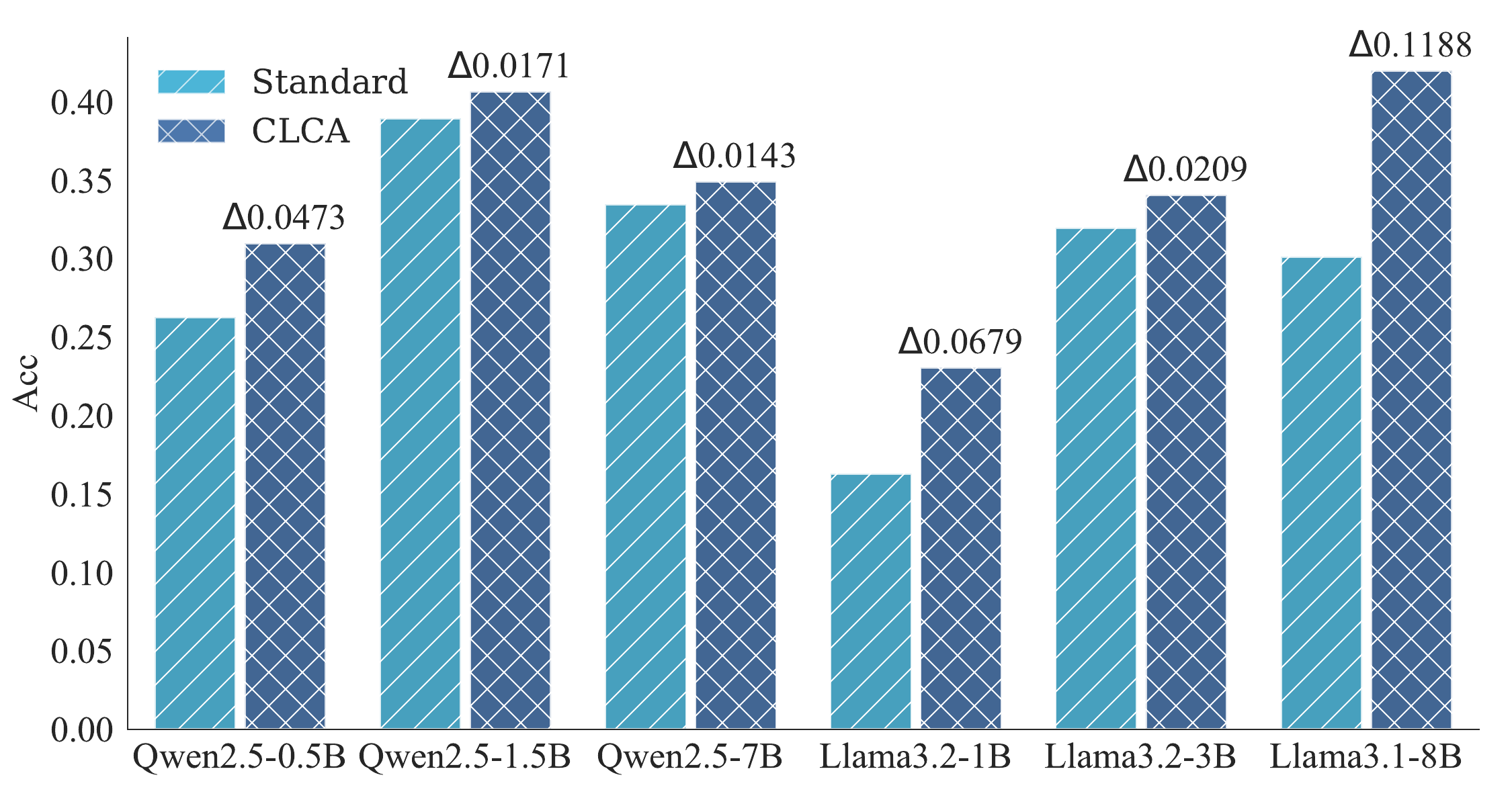}
        \caption{Individual-level accuracy (higher is better) between the model prediction and WVS data.}
        \label{fig:llama_xling_acc}
    \end{subfigure}
    \caption{Average performance of models (\texttt{Standard} is the zero-shot evaluation of the persona baseline described in \S\ref{subsec:method}., \methodname{} is the adaptation in English) responding to survey questions in the native language of the culture. Results are averaged over all languages.}
    \label{fig:xling}
\end{figure*}

\subsection{Intent Understanding is Important in \methodname{}}

Our main results in Table \ref{tab:main} and experiments in the previous subsections show that training on social data is important and effective for culture adaptation. Here, we further analyze the significance of intent understanding in this adaptation process. We perform experiments with 1) training on the conversation data only (i.e., \texttt{dialogue\_only}); and 2) training with intent understanding with respect to social and cultural norms (i.e., \texttt{intent\_only}). The results are in Table \ref{tab:ablation_intent}.

We observe that training on the conversation data alone improves individual-level accuracy by 2.91\% points and improves KL-D by 0.0307. While it is interesting to see that training with intent alone has nearly no effect on the results, it can further improve the individual-level accuracy by 5.2\% points from conversation training. Similar compounding effects are also observed for Qwen models in Table \ref{tab:ablation_intent_qwen} (in Appendix). This confirms that the combination of two cultural learning strategies (i.e., imitative, instructed and intent) is more effective. 

\begin{table}[h!]
    \centering
\resizebox{0.92\linewidth}{!}{
    \begin{tabular}{l cc}
\toprule
Model & Acc \textcolor{SeaGreen}{$\uparrow$} & KL-D \textcolor{SeaGreen}{$\downarrow$}\\
\midrule
    Llama3.1 8B  & 0.3162 &  0.6011 \\
    Llama3.1 8B $_{\methodname{}}$ & \textbf{0.3973} & \textbf{0.5112}\\ 
    \hdashline
    Llama3.1 8B $_{\methodname{} \texttt{ intent\_only}}$ & 0.3117 & 0.6037 \\
    Llama3.1 8B $_{\methodname{} \texttt{ dialogue\_only}}$   & 0.3453 &  0.5704\\ 
    \bottomrule
    \end{tabular}
    }
    \caption{Ablation study of the Llama3.1 8B model: training on conversation only, intent understanding only, versus both objectives combined (i.e., \methodname{}). The best results are bolded, and the direction of the arrows indicates if the metrics should be maximized or minimized.}
    \label{tab:ablation_intent}
\end{table}

\subsection{Zero-shot Value Transfer to Other Languages}\label{sec:xling}

So far, we have used English data to improve the cultural value alignment of LLMs, with evaluations conducted in English. Next, we evaluate the Llama 3.1 8B model (selected for its significant improvements after adaptation and exceptional task performance) using translated WVS questions in the respective languages of the target cultures. British culture is excluded as its primary language, English, requires no translation. Survey questions and prompt templates are translated using GPT-4.

Figure \ref{fig:xling} presents the results for the six multilingual models, averaged across languages. Overall, the models show consistent improvements in both culture-level KL-D and individual-level accuracy. Notably, the Llama models exhibit greater improvements compared to the Qwen models, although they are initially less aligned with respected cultural values. It is also interesting to observe that while Qwen2.5 7B shows no improvement in English evaluations (Table \ref{tab:main}), it demonstrates improved performance in multilingual evaluations, with a 1.43\% increase in individual-level accuracy and a reduction of 0.0145 in KL-D.

\subsection{Data Generation Model}
Another key question is whether the adaptation works only with the Llama3.1 70B model as a teacher.
To assess the generalizability of our findings, we use the same pipeline to collect simulated data from the Qwen2.5 32B model. 
This data was then used to train the Llama3.1 8B model, resulting in an average KL-D of 0.5617 and an accuracy of 0.3487. Although these results outperform the baselines, they fall short of those achieved using data generated by the Llama3.1 70B model. The discrepancy stems from two factors: a smaller training dataset after filtering and the quality of the generated content, including issues like code-mixing in conversations. 
While the teacher model's capability and the quality of generated data influence adaptation results, the improvements highlight cultural learning as an effective adaptation strategy.\footnote{Appendix \ref{app:ablation} presents additional ablation studies including training without data filtering and anthropological prompting, further highlighting the potential of cultural learning.}

\section{Conclusion}

In this work, we investigate the effectiveness of cultural learning-based training for cultural value adaptation in LLMs. We propose a novel framework, \methodname{}, that leverages culturally adapted social scenarios, social interactions, intents and their relation to social and cultural norms.

We validate the effectiveness of \methodname{}, showcasing how LLMs can be adapted to align with various cultural values across different model architectures and sizes. 
It provides early evidence that social interaction data can help align cultural values.
Our analysis reveals the importance of intent understanding and a complementary relationship between the two cultural learning strategies. Our findings highlight cultural learning as a promising direction for adaptation, paving the way toward building more inclusive and culturally aware NLP.

\section*{Limitations}

There are several limitations to our work:

\textbf{Bias in synthetic data generation and LLM-as-a-Judge.} 
In our experiments, we use LLMs to role-play individuals from different cultures. While training on this synthetic data improves alignment with human survey responses on cultural values, the data could reflect biases, stereotypes, or unrealistic interactions and caricatures associated with cultural groups \cite{caricature/emnlp/cheng-etal-2023-compost, Wang2024LargeLM} due to their synthetic nature. While beyond our scope, we provide qualitative studies into the data which highlight the need for further research into this area (Appendix \ref{app:data_quality}).

Additionally, our data collection is conducted in English rather than multilingually. Collecting multilingual data would require the model to demonstrate greater fluency and authenticity in generating conversations in different social settings. This ability is often overlooked in current LLM evaluations and culturally aware NLP \cite{culture_survey}, which primarily focuses on multiple-choice questions or reasoning tasks. Addressing this gap is a goal for future work but lies beyond the scope of this paper.

Finally, we employ the LLM-as-a-Judge for data filtering, which has become a common practice \cite[inter alia]{llm/nips/Ouyang0JAWMZASR22,nips/chatbotarena/ZhengC00WZL0LXZ23,reward/dang-etal-2024-rlhf, iclr/prometheus/KimS0JLLYSKTS24} in NLP. Although model-based judgments correlate with human evaluations, they still exhibit discrepancies, indicating potential biases that require further investigation, especially in diverse cultural contexts.

\textbf{Real social interaction conversations.} While our proposed cultural learning-based framework has demonstrated effectiveness, its robustness in real-world scenarios remains uncertain. In this paper, we demonstrate that a hypothetical culture expert model (e.g., Llama3.1 70B, the data generation model), can improve weaker models aligning to cultural values. Since individuals from the target culture are the ultimate cultural experts, incorporating real human interactions into cultural learning-based training presents an exciting opportunity for improvement. However, their effectiveness remains unknown and requires further investigation.

\textbf{Low-resource cultures.} Our paper takes an exciting first step toward exploring whether a theory-based approach, cultural learning, can be effectively used for cultural value adaptation. We focused on more widely available cultures to validate our idea and leave the important question of low-resource cultures for future work. In this study, we selected a diverse range of cultures based on the availability of sufficient responses from the WVS, which we believe provides adequate validation for our proposed learning method. To address challenges related to low-resource cultures with cultural learning-based methods, a potential direction is to collect more real human data. 

\textbf{Survey evaluation as a proxy.} In this study, we evaluate the adaptation results using WVS data. While WVS data serves as a proxy \cite{DBLP:journals/corr/abs-2403-15412} for human values, it has limitations, such as survey sample size and potential gaps between survey responses and actual values. In future work, we aim to incorporate a broader range of proxies and downstream tasks to enable a more comprehensive evaluation.

\section*{Ethics Statement}
In this work, we aim to investigate the effectiveness of cultural learning-based training strategies for adapting LLMs to different cultural values. Our primary goal is not to treat models as potential human subjects or anthropomorphize LLMs. We strive to address technical challenges responsibly, and we encourage users of our findings to adhere to ethical and moral guidelines.

Through this research, we demonstrate the potential of a human-inspired methodology to improve LLMs for different cultures. We seek to inspire interdisciplinary collaborations to ethically design technology that meets human needs, advancing NLP that promotes respect for cultural variations globally.

\section*{Acknowledgements}
This work has been funded by the LOEWE Distinguished Chair ``Ubiquitous Knowledge Processing'', LOEWE initiative, Hesse, Germany (Grant Number: LOEWE/4a//519/05/00.002(0002)/81). This work has also been supported by the UK Research and Innovation (UKRI) Frontier Research Grant EP/Y031350/1 EQUATE (the UK government’s funding guarantee for ERC Advanced Grants) awarded to Anna Korhonen at the University of Cambridge. 

We thank Thy Thy Tran, Sheng Lu, and Fengyu Cai for their feedback on a draft of this work.

\bibliography{anthology, culture_revised, cl}
\bibliographystyle{acl_natbib}

\appendix

\clearpage

\begin{figure}
    \centering
    \includegraphics[width=\linewidth]{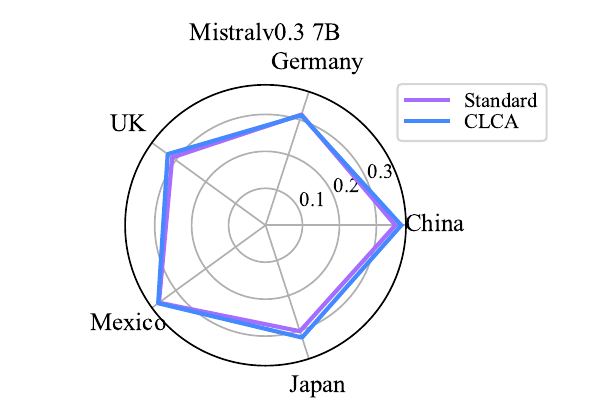}
    \caption{Individual-level accuracy for Mistral model.}
    \label{fig:mistral}
\end{figure}

\begin{table}[]
\centering
\resizebox{0.76\columnwidth}{!}{
\begin{tabular}{l c}
\toprule
 &  Avg. \texttt{Acc} \textcolor{SeaGreen}{$\uparrow$} \\
\midrule
Llama3.1 8B & 0.3162 \\
Llama3.1 8B$_{\methodname{}}$ & \textbf{0.3973}  \\
\hdashline
Llama3.2 3B &  0.2983  \\
Llama3.2 3B$_{\methodname{}}$ & \textbf{0.3148} \\
\hdashline
Llama3.2 1B & 0.3275 \\
Llama3.2 1B$_{\methodname{}}$  & \textbf{0.3293} \\ 
\midrule
Qwen2.5 7B &  \textbf{0.4412} \\
Qwen2.5 7B$_{\methodname{}}$ & 0.4337 \\
\hdashline
Qwen2.5 1.5B & 0.3211 \\
Qwen2.5 1.5B$_{\methodname{}}$ & \textbf{0.3645} \\
\hdashline
Qwen2.5 0.5B & 0.3272 \\
Qwen2.5 0.5B$_{\methodname{}}$ & \textbf{0.3698} \\
\midrule
Mistral-v0.3 7B & 0.3273 \\
Mistral-v0.3 7B$_{\methodname{}}$ & \textbf{0.3372}\\
\bottomrule
\end{tabular}}
\caption{Individual-level accuracy averaged across cultures.}
\label{tab:acc_main}
\end{table}

\begin{table}[h!]
    \centering
\resizebox{0.98\linewidth}{!}{
    \begin{tabular}{l cc}
\toprule
    Model & Acc \textcolor{SeaGreen}{$\uparrow$} & KL-D \textcolor{SeaGreen}{$\downarrow$}\\
    \midrule
    Qwen2.5 1.5B  & 0.3211 &  0.5956 \\
    Qwen2.5 1.5B $_{\methodname{}}$ & \textbf{0.3645} & \textbf{0.5522}\\ 
    \hdashline
    Qwen2.5 1.5B $_{\methodname{}\texttt{ intent\_only}}$ & 0.3084 & 0.6108 \\
    Qwen2.5 1.5B $_{\methodname{}\texttt{ dialogue\_only}}$ & 0.3184 & 0.5962\\ 
    \midrule
    Qwen2.5 0.5B & 0.3272 & 0.5748\\
    Qwen2.5 0.5B$_{\methodname{}}$ & \textbf{0.3698} &  \textbf{0.5317} \\
    \hdashline
    Qwen2.5 0.5B $_{\methodname{}\texttt{ intent\_only}}$ & 0.3292 & 0.5726 \\
    Qwen2.5 0.5B $_{\methodname{}\texttt{ dialogue\_only}}$ & 0.3598 & 0.5499\\ 

    \midrule
    Llama3.2 3B &  0.2983 & 0.6319 \\
    Llama3.2 3B$_{\methodname{}}$ & \textbf{0.3148} & \textbf{0.6078}\\
    \hdashline
    Llama3.2 3B $_{\methodname{}\texttt{ intent\_only}}$ & 0.2969 & 0.6336 \\
    Llama3.2 3B $_{\methodname{}\texttt{ dialogue\_only}}$ & 0.3058 & 0.6204\\ 

    \midrule
    Llama3.2 1B & 0.3275 & 0.6081 \\
    Llama3.2 1B$_{\methodname{}}$  & \textbf{0.3293} & \textbf{0.5931} \\ 
    \hdashline
    Llama3.2 1B $_{\methodname{}\texttt{ intent\_only}}$ & 0.3265 & 0.6092 \\
    Llama3.2 1B $_{\methodname{}\texttt{ dialogue\_only}}$ & 0.3208 & 0.6064\\ 
    \bottomrule
    \end{tabular}
    }
    \caption{Additional ablation results for other Llama and Qwen models: training on conversation only, intent understanding only, versus both objectives combined (i.e., \methodname{}). The best results are bolded, and the direction of the arrows indicates if the metrics should be maximized or minimized. In general, training with intent only does not improve results. However, combining both approaches yields significant improvements.}
    \label{tab:ablation_intent_qwen}
\end{table}

\begin{figure*}
    \centering
    \includegraphics[width=\linewidth]{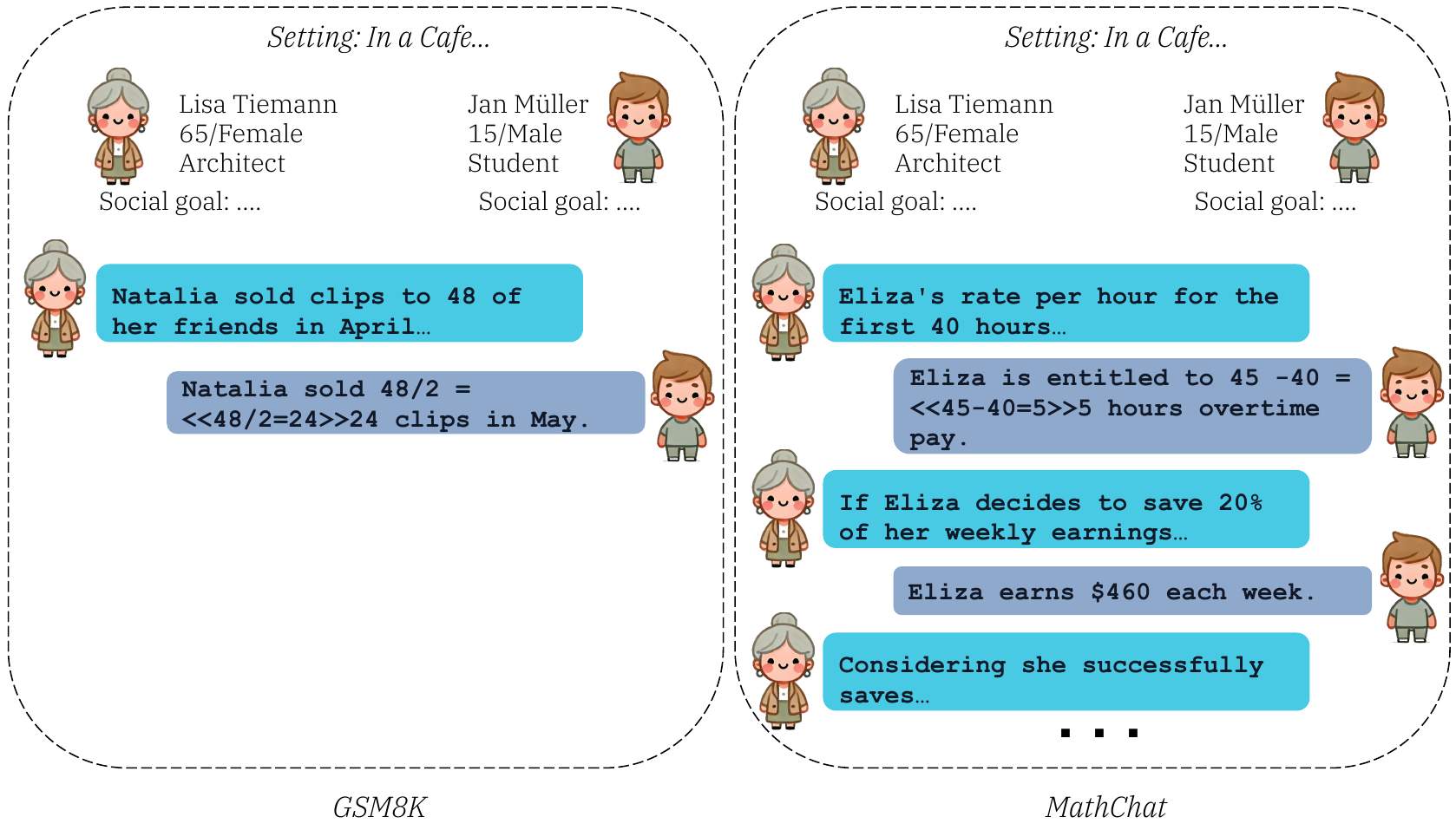}
    \caption{Illustration of training with GSM8K and training with MathChat (the follow-up setting). In these two experiments, we keep the social setting, participants and their social goals the same as \methodname{} training, while conversations are replaced with GSM8K or MatchChat which reflects minimal social and cultural information. The example (including the setting) is shortened for illustration purposes.}
    \label{fig:gsm8k}
\end{figure*}

\begin{table}[]
\centering
\resizebox{\columnwidth}{!}{
\begin{tabular}{p{0.9\columnwidth}} 
    \toprule
    Example\\
    \midrule
    Setting: At a bustling shopping mall a tourist is trying to find the nearest restroom.\\
    Li Wei: 35 / Male, Mall Security Guard \\
    Goal: To assist Zhang in finding the restroom while maintaining the flow of traffic in the mall. \\
    Zhang Qi: 45 / Male, Tourist \\
    Goal: To find the nearest restroom as quickly as possible.\\
    \hdashline
    \hdashline

    Li Wei: Excuse me, sir. Are you looking for something? \\
    \textit{Intent:} Offering help. \\
    Zhang Qi: Ah, restroom. I'm looking for the restroom. Could you tell me where it is? \\
    \textit{Intent:} To get directions to the restroom. \\
    
    \hdashline
    Li Wei: The restroom is just down that hallway, sir. You can't miss it. \\
    \textit{Intent:} Li Wei's intent is to politely and efficiently provide directions to the tourist. \\
    Zhang Qi: Thank you, I'll just go take a look. \\
    \textit{Intent:} Zhang Qi intends to politely thank Li Wei and follow his directions to find the restroom. \\
    
    \hdashline
    Li Wei: Sir, would you like me to escort you to the restroom? \\
    \textit{Intent:} Li Wei is showing respect and courtesy, as is customary in Chinese culture, especially when interacting with an older person (Zhang Qi is years older than Li Wei). \\
    Zhang Qi: Thank you for your help, I'll just go now. \\
    \textit{Intent:} Politeness and appreciation. \\

    \bottomrule
\end{tabular}}
    \caption{An generated conversation example with intents.}
    \label{tab:example}
\end{table}

\begin{table}[]
    \centering
    \resizebox{0.98\linewidth}{!}{
    \begin{tabular}{l}
    \toprule
    Categories \\
    \midrule
        Ethical Values \\
        Perceptions of Security\\
        Religious Values\\
        Happiness and Wellbeing\\
        Demographic and Socioeconomic Variables\\
        Perceptions about Science and Technology\\
        Social Capital, Trust and Organizational Membership\\
        Political Interest and Political Participation\\
        Perceptions of Corruption\\
        Perceptions of Migration\\
        Social Values, Norms, Stereotypes\\
        Political Culture and Political Regimes\\
        Economic Values\\
    \bottomrule
    \end{tabular}}
    \caption{All Question categories in the World Value Survey.}
    \label{tab:wvs_cat}
\end{table}

\begin{table}[]
    \centering
\resizebox{0.99\linewidth}{!}{
    \begin{tabular}{lccccc}
    \toprule
    Culture & Scenarios & Size & AT & AW & CI\\
    \midrule
        China  &225 & 107 & 6.37 & 77.45 & 45.38 \\
        Germany   &208 & 85 & 6.92 & 76.42 & 31.87 \\
        UK  &193 & 143 & 7.04 & 75.48 & 29.52\\
        Mexico  &221 & 105  & 6.10 & 79.14 & 53.21\\
        Japan &209 & 69 & 5.36 & 74.74 & 33.30\\
    \bottomrule
    \end{tabular}
    }
    \caption{Data statistics of the number of social scenarios, number of conversations after filtering, average turns (\textbf{AT}), average words per turn (\textbf{AW}) and percentage of intents with cultural context (\textbf{CI}) in the dataset.}
    \label{tab:data_stats}
\end{table}

\section{Additional Ablations}\label{app:ablation}

\noindent{\textbf{No Data Filtering.}} Prior work shows that data filtering is important to achieve better performance of synthetic data. Here, we ablate the effect of data filtering with Llama3.1 8B model, and the results are in Table \ref{tab:filter}. While showing improvements after training, this shows that having quality data is important.

\begin{table}[]
    \centering
    \resizebox{0.95\linewidth}{!}{
    \begin{tabular}{l cc}
\toprule
Model & Acc \textcolor{SeaGreen}{$\uparrow$} & KL-D \textcolor{SeaGreen}{$\downarrow$}\\
\midrule
    Llama3.1 8B & 0.3162 &  0.6011\\
    Llama3.1 8B$_{\methodname{} \texttt{ no\_filter}}$ & 0.3608 & 0.5639\\
    Llama3.1 8B$_{\methodname{}}$ & \textbf{0.3973} & \textbf{0.5112}\\
    \bottomrule
    \end{tabular}}
    \caption{Ablation results using unfiltered data versus data with filtering on Llama3.1 8B.}
    \label{tab:filter}
\end{table}

\noindent{\textbf{Prompting.}} We additionally experimented with Anthropological prompting \cite[\texttt{anthropological}]{nvm/anthroprompt/acl/abs-2402-13231} for Llama3.1 8B, Qwen2.5 7B and Mistral-v0.3 7B models. This method uses personas along with an anthropological reasoning guidance prompt to elicit the LLM's explanation before answering survey questions. Note that the evaluation time for anthropological prompting per persona is significantly longer than other evaluation methods, as it requires extended reasoning generation prior to answering. Therefore, we allocate a fixed evaluation time budget using anthropological prompting: 6 hours per culture (30 hours in total on a single A6000 GPU, 4-bit inference, 50 personas) using the Llama3.1 8B model, nearly double the time used in other evaluations (e.g., 3 to 4 hours per culture, 4-bit inference) of the same model per culture. 

The evaluation results are shown in Table \ref{tab:anth}, along with cultural prompting and the persona baseline. Overall, the performance of anthropological prompting is relatively inconsistent compared to the persona baseline or cultural prompting. Interestingly, anthropological prompting achieves better KL-D but worse individual-level accuracy for Llama3.1 8B, while other prompting methods are more stable across models and achieve better results. Nonetheless, existing prompting methods perform worse than training using \methodname{} in general (as seen in our main paper, Table \ref{tab:main}).

\begin{table}[]
    \centering
    \resizebox{0.98\linewidth}{!}{
    \begin{tabular}{l cc}
    \toprule
    Model & Acc \textcolor{SeaGreen}{$\uparrow$} & KL-D \textcolor{SeaGreen}{$\downarrow$}\\
    \midrule
    Llama3.1 8B & 0.3162 &  0.6011\\
    Llama3.1 8B \texttt{cultural} & 0.3274 & 0.6020\\
    Llama3.1 8B \texttt{anthropological} & 0.3039 & 0.5694\\
    \hdashline
    Qwen2.5 7B & \textbf{0.4412} &  \textbf{0.4817} \\
    Qwen2.5 7B \texttt{cultural} &  0.3921 &  0.5364 \\
    Qwen2.5 7B \texttt{anthropological} &  0.3420 & 0.5561 \\
    \hdashline
    Mistral-v0.3 7B & 0.3273 & 0.6081 \\
    Mistral-v0.3 7B \texttt{cultural} & 0.3101 & 0.6227 \\
    Mistral-v0.3 7B \texttt{anthropological} &  0.2255 & 0.6604 \\
    \bottomrule
    \end{tabular}}
    \caption{Results using different prompting methods on Llama3.1 8B, Qwen2.5 7B and Mistral-v0.3 7B.}
    \label{tab:anth}
\end{table}

\noindent\textbf{More Ablations Using MathChat.} 
The average number of turns in MathChat (3.66 turns) is approximately half of the generated social interaction dialogues (Table \ref{tab:data_stats}). To investigate this further, we perform an additional ablation experiment by concatenating two randomly chosen MathChat dialogues for training (\texttt{MathChat\_Long}). The results in Table \ref{tab:ablation_gsm8k2} show that incorporating \texttt{MathChat\_Long} does not impact the model's performance, indicating that the number of turns does not influence the training results here.

\begin{table}[]
    \centering
    \resizebox{0.92\linewidth}{!}{
    \begin{tabular}{l cc}
    \toprule
    Model & Acc \textcolor{SeaGreen}{$\uparrow$} & KL-D \textcolor{SeaGreen}{$\downarrow$}\\
    \midrule
    Llama3.1 8B & 0.3162 &  0.6011\\
    Llama3.1 8B$_{\methodname{}}$ & \textbf{0.3973} & \textbf{0.5112}\\
    \hdashline
    Llama3.1 8B$_\texttt{GSM8K}$  & 0.3287 & 0.5902\\
    Llama3.1 8B$_\texttt{MathChat}$  & 0.3260 & 0.5818\\
    Llama3.1 8B$_\texttt{MathChat\_Long}$  & 0.3156 & 0.6041\\
    \hdashline
    Llama3.1 8B$_\texttt{Wiki}$  & 0.3238	&0.6010 \\
    Llama3.1 8B$_\texttt{CK\_Roleplaying}$  & 0.3151 & 0.6130 \\
    \bottomrule
    \end{tabular}}
    \caption{Comparison of Llama3.1 8B model trained with a reasoning-only dataset, cultural knowledge-only datasets versus training with social conversation. All models are instruction-tuned, the direction of the arrows indicates if the values should be maximized or minimized.}
    \label{tab:ablation_gsm8k2}
\end{table}

\noindent\textbf{Ablations Using Cultural Knowledge.} 
As the prior experiment has shown, reasoning data does not improve the models' value alignment. Here, we investigate whether cultural knowledge helps with value alignment. To the best of our knowledge, there is no existing dataset containing cultural knowledge in a conversational format without social interactions. Therefore, we perform two additional ablations with synthetic data as follows.

The first experiment (\texttt{Wiki}) uses Wikipedia pages that provide high-level descriptions of a culture. We prompt the GPT-4 model to generate factual conversation grounded in the provided paragraphs (3 consecutive paragraphs randomly sampled every time) from selected Wikipedia pages (in Table \ref{tab:url}). Our goal is to eliminate cultural knowledge as a contributing factor in value adaptation. We generated 200 conversations and trained the model using the same settings as in the GSM8K and MathChat experiments.

The second experiment (\texttt{CK\_Roleplaying}) utilizes cultural concepts sourced from Wikipedia (e.g., Heinerfest or Kung Pao Chicken), covering topics like food, holidays, dances, and music. We then apply the same data generation pipeline as \methodname{}, using the Llama 3.1 70B model.
All social settings and goals from the filtered data in \methodname{} are replaced with \textit{\{Concept\_Name\}} and \textit{Want to share factual knowledge about \{Concept\_Name\} from \{culture\} culture}. We generate role-playing conversations while keeping them focused on cultural concepts without exhibiting social interactions. The objective is to eliminate the possibility that linguistic or stylistic cues from role-playing influence value adaptation, ensuring that value alignment primarily results from social interactions. For each pair of characters used in training, we generate two conversations.

The results in Table \ref{tab:ablation_gsm8k2} (last two rows) show that cultural knowledge alone does not impact the model's performance.

\begin{table}[]
    \centering
    \begin{tabular}{c}
    \toprule
         Title  \\
    \midrule
    \href{https://en.wikipedia.org/wiki/Culture_of_the_United_Kingdom}{Culture of the United Kingdom} \\
    \href{https://en.wikipedia.org/wiki/Culture_of_Germany}{Culture of Germany} \\
    \href{https://en.wikipedia.org/wiki/Chinese_culture}{Chinese culture} \\
    \href{https://en.wikipedia.org/wiki/Culture_of_Mexico}{Culture of Mexico}\\
    \href{https://en.wikipedia.org/wiki/Culture_of_Japan}{Culture of Japan}\\
    \bottomrule
    \end{tabular}
    \caption{Titles of the Wikipedia pages used for data generation.}
    \label{tab:url}
\end{table}

\section{Hyperparameters and Hardware} \label{app:hparams}

In our experiments, we use the following hyperparameters for models. We perform the hyperparameter search with learning rate over [1e-5, 5e-5, 1e-4], training epochs of 1 or 3. Table \ref{tab:hparams} outlines all the hyperparameters. 

The experiments were conducted on a server with a single NVIDIA A6000 or A100 GPU, depending on availability. Inference was performed in 4-bit precision. For the 7B and 8B models, the inference time ranged from 3 to 4 hours per culture.

\begin{table}[]
    \centering
\resizebox{0.98\linewidth}{!}{

    \begin{tabular}{lc}
    \toprule
    Parameter & Value \\
    \midrule
      Batch Size   &  8 \\
      Learning Rate &  Llama=1e-4, Qwen=1e-4, Mistral=5e-5 \\
      Epochs & Llama=3, Qwen=1, Mistral=3\\
      LoRA r & 4\\
      LoRA alpha & 0.1 \\
      LoRA dropout & 0.5 \\
      LoRA target modules & q\_proj, v\_proj \\
    \bottomrule
    \end{tabular}}
    \caption{Hyperparameters used in our experiments.}
    \label{tab:hparams}
\end{table}

\begin{table}[h!]
\centering
\resizebox{0.78\linewidth}{!}{
\begin{tabular}{l cc}
\toprule
& Avg. \texttt{JS-D} \textcolor{SeaGreen}{$\downarrow$} \\
\midrule
Llama3.1 8B &  0.5134  \\
Llama3.1 8B$_{\methodname{}}$ & \textbf{0.4303}  \\
\hdashline

Llama3.2 3B &   0.5626 \\
Llama3.2 3B$_{\methodname{}}$ & \textbf{0.5402} \\
\hdashline
Llama3.2 1B &  0.5592  \\
Llama3.2 1B$_{\methodname{}}$  &  \textbf{0.5195}\\ 

\midrule
Qwen2.5 7B &  \textbf{0.4267} \\
Qwen2.5 7B$_{\methodname{}}$ & 0.4279\\
\hdashline
Qwen2.5 1.5B &  0.5138\\
Qwen2.5 1.5B$_{\methodname{}}$ & \textbf{0.4817} \\

\hdashline
Qwen2.5 0.5B &  0.4575 \\
Qwen2.5 0.5B$_{\methodname{}}$ & \textbf{0.4100} \\

\midrule
Mistral-v0.3 7B  & 0.5604 \\
Mistral-v0.3 7B$_{\methodname{}}$ & \textbf{0.5522}\\

\bottomrule
\end{tabular}}
\caption{The JS-D between the distribution of predicted answers and the distribution of the ground truth answers from the WVS survey of various models on different cultures. All models are instruction-tuned, the green arrow indicates the lower the JS-D the better, and the bold indicates the better result.}
\label{tab:main_jsd}
\end{table}

\section{Alternative Metrics}\label{app:jsd}

In our main paper, we use KL-D to measure the similarity between predicted answers to the ``ground truth'' human answer distributions. This is used since our goal is to achieve distributional similarity using the approximate distributions (i.e., answers from LLMs) to real distributions (i.e., answers from humans). 

Alternatively, a symmetric metric, Jensen-Shannon Distance (JS-D), as used in \citet{GlobalOpinionQA/corr/abs-2306-16388} can be used. JS-D is defined as: 
\begin{align*}
&D_{JS}(P_i; Q_i) \\
&= \sqrt{\frac{1}{2} D_{KL}(P_i; m_i) + \frac{1}{2} D_{KL}(Q_i; m_i)} ,
\end{align*}
\noindent
where $m_i$ is the pointwise mean of $P_i$ and $Q_i$, and $D_{KL}(P_i; m_i)$ is the KL-D for question $i$ from the model, $D_{KL}(Q_i; m_i)$ is the KL-D for question $i$ from the survey. The final $D_{JS}$ is averaged over all questions. When the distributions are similar, the JS-D value is smaller.

The results of the persona baseline and \methodname{} presented in Table \ref{tab:main} of our main paper, using JS-D, are provided in Table \ref{tab:main_jsd}. Since JS-D is derived from KL-D, the results exhibit similar trends. 
\methodname{} enhances the alignment of cultural values across models of various sizes, with the Qwen2.5 7B model being an outlier.

\section{Synthetic Data Quality}\label{app:data_quality}

In this work, we rely on model filtering as an approximation for quality. In addition, we provide qualitative studies on the overall conversation's cultural acceptability and intent acceptability. 

We recruit participants from Prolific based on nationality and language proficiency to approximate cultural backgrounds. We also require English proficiency, as our synthetic data is in English.

\noindent\textbf{Intents.} We randomly sampled 5 conversations per culture (total of 320 intents) that passed the filter and performed the human evaluation of the intents with two annotators from each culture. 
We asked the annotators to assess the plausibility of the general and cultural intents, aggregating the results using a majority vote. The overall evaluation results are in Table~\ref{tab:intent_scores}. The intents have an overall acceptability rate of 86.82\% on average across cultures. However, this value drops to 78.70\% for the cultural intents, which we still consider acceptable.

\begin{table}[]
    \centering
    \resizebox{0.8\columnwidth}{!}{
    \begin{tabular}{lcc}
        \toprule
        Culture & Intent & Cultural Intent \\
        \midrule
        Germany & 0.7424 & 0.6094 \\
        Mexico & 0.8305 & 0.7143 \\
        Japan & 0.9661 & 0.9200 \\
        UK & 0.8592 & 0.8868 \\
        China & 0.8438 & 0.7500 \\
        \bottomrule
    \end{tabular}
    }
    \caption{Intent and cultural intent evaluations.}
    \label{tab:intent_scores}
\end{table}

\noindent\textbf{Conversations.} We randomly sampled five conversations per culture and asked human evaluators from each culture to assess and provide feedback on the data's acceptability with respect to their cultural norms. Overall, participants rated the Chinese and Japanese conversations as acceptable to excellent (5 out of 5). In contrast, this rating dropped for German, British and Mexican cultures (4 out of 5). 
While this small-scale qualitative study cannot determine whether the synthetic data truly aligns with cultural aspects, the results indicate that it captures \textit{some cultural nuances}, supporting its use in our cultural learning-based training in this work.

However, our study revealed significant subjectivity, where it is possible for human evaluators to assign opposite labels to the same data (e.g., excellent example versus impossible for the culture). Additionally, an evaluator noted that while the data represent cultural aspects, their assessment reflects only the perspective of their specific region. 

This highlights the need for carefully designed, large-scale studies across a broad range of demographic groups, improved role-playing methods for individuals from different cultures, and rigorous metrics to evaluate generational, behavioural alignment with a culture.

\section{Additional Cultural Information to Guide the Conversation Generation}\label{app:cultural_guide}

We incorporate additional cultural information to guide the role-playing per culture. We supplement the system prompt with information from Hofstede's cultural dimensions \cite{hofstede2010cultures} and Inglehart–Welzel cultural map \cite{inglehart2005christian}.

We map Hofstede's cultural dimensions values \cite{hofstede_cd, culturefactor} for the respective cultures into verbal descriptions such as ``highly hierarchical'', ``moderately collective'' etc. The Hofstede cultural dimensions consist of six dimensions, including:
\begin{itemize}[noitemsep, topsep=0.1pt]
    \item  Power distance (verbalized as hierarchical versus equal)
    \item  Individualism / Collectivism (verbalized as individualistic versus collective)
    \item Motivation towards achievement and success (verbalized as motivation for achievement and success)
    \item Uncertainty avoidance (verbalized as risk-taking versus uncertainty avoidance)
    \item Long-term orientation / Short-term orientation (verbalized as normative versus pragmatic)
    \item Indulgence / Restraint (verbalized as restrained versus indulgent)
\end{itemize}

The resulting verbalized descriptions of Hofstede's cultural dimensions values are in Table \ref{tab:hofstede}. 

\begin{table}[]
    \centering
    \resizebox{\columnwidth}{!}{
    \begin{tabular}{l p{0.8\columnwidth}}
    \toprule
       Culture  & Dimensions \\
    \midrule
        China & highly hierarchical, moderately collective, moderate motivation for achievement and success, moderately risk-taking, highly pragmatic, highly restrained \\
        Mexico & highly hierarchical, moderately collective, moderate motivation for achievement and success, high uncertainty avoidance, highly normative, highly indulgent \\
        Japan & moderately hierarchical, moderately individualistic, high motivation for achievement and success, high uncertainty avoidance, highly pragmatic, moderately restrained \\
        Germany & moderately equal, highly individualistic, moderate motivation for achievement and success, moderately uncertainty avoidance, moderately pragmatic, moderately restrained \\
        British & moderately equal, highly individualistic, moderate motivation for achievement and success, moderately risk-taking, moderately pragmatic, moderately indulgent \\
    \bottomrule
    \end{tabular}
    }
    \caption{Mapping of Hofstede's cultural dimensions to verbalized form for prompting.}
    \label{tab:hofstede}
\end{table}

The Inglehart–Welzel cultural map consists of two dimensions\footnote{\url{https://www.worldvaluessurvey.org/WVSContents.jsp}}, including:

\begin{itemize}[noitemsep, topsep=0.1pt]
    \item Traditional values versus secular values (verbalized as traditional versus secular)
    \item Survival values versus self-expression values (verbalized as-is)
\end{itemize}

Similarly, we verbalize the cultural dimensions, which are in Table \ref{tab:inglehart}.

\begin{table}[]
    \centering
    \resizebox{\columnwidth}{!}{
    \begin{tabular}{l p{0.8\columnwidth}}
    \toprule
       Culture  & Dimensions \\
    \midrule
        China & little survival values, moderately secular \\
        Mexico & little self-expression values, moderately traditional \\
        Japan & moderate self-expression values, uttermost secular \\
        Germany & uttermost self-expression values, moderately secular \\
        British & uttermost self-expression values, moderately secular \\
    \bottomrule
    \end{tabular}
    }
    \caption{Mapping of Inglehart–Welzel cultural map to verbalized form for prompting.}
    \label{tab:inglehart}
\end{table}

\section{Prompts}\label{app:prompts}

Table \ref{tab:cult_adapt_prompt} to Table \ref{tab:filter_prompt} provide the prompts used in our experiments. 

\begin{table}[]
\centering
\resizebox{\columnwidth}{!}{
\begin{tabular}{p{0.9\columnwidth}} 
    \toprule
    Culture Adaptation (User) Prompt\\
    \midrule
    Here is a generic social interaction.
    Could you update the name, occupation, setting, goals, and secrets to make the information logically coherent so that it represents a believable scenario that could occur within \texttt{\{culture\}}?
    Please change all participants' names to diverse \texttt{\{culture\}} names with diverse occupations. The goals and secrets should be relevant to the interaction setting and play a key role in inciting or resolving conflicts in this interaction. Return the data using the same JSON schema in English without any explanation.\\
    Input: \texttt{\{scenario\}} \\
    Output:\\
    \bottomrule
\end{tabular}}
    \caption{Prompt used to create culturally adapted scenarios.}
    \label{tab:cult_adapt_prompt}
\end{table}

\begin{table}
\centering
\resizebox{\columnwidth}{!}{
\begin{tabular}{p{0.9\columnwidth}} 
\toprule
Persona (System) Prompt\\
\midrule
You are currently living in \texttt{\{country\}} \\
and here is your basic demographic information: \\
Settlement: \texttt{\{settlement\}}, \texttt{\{region\}} \\
Gender: \texttt{\{gender\}} \\
Age: \texttt{\{age\}} \\
Born in \texttt{\{country\}}: \texttt{\{born\}} \\
Marital status: \texttt{\{marital\_status\}} \\
Number of people in household: \texttt{\{household\}} \\
Education: \texttt{\{education\}} \\
Profession: \texttt{\{profession\}} \\
Employment: \texttt{\{employeed\}} \\
Class: \texttt{\{classes\}} \\
\bottomrule
\end{tabular}}
\caption{Demographic information used in our work for evaluation.}
\label{tab:demo}
\end{table}

\begin{table}[]
\centering
\resizebox{\columnwidth}{!}{
\begin{tabular}{p{0.9\columnwidth}} 
    \toprule
    Evaluation Question (User) Prompt\\
    \midrule
    Please answer the following question, output the integer option when instructed, don't explain: \\
    QUESTION: \texttt{\{question\}}\\
    ANSWER: \\
    \bottomrule
\end{tabular}}
    \caption{Prompt used to evaluate WVS questions.}
    \label{tab:user_prompt}
\end{table}

\begin{table}[ht]
\centering
\resizebox{\columnwidth}{!}{
\begin{tabular}{p{0.9\columnwidth}} 
\toprule
Intent Generation Prompts\\
\midrule
System Prompt \\
\hdashline
Here is the basic information about this conversation.\\
Scenario:  \texttt{\{setting\}}\\
Information about  \texttt{\{name\}}: \\
    Background:  \texttt{\{background\}} \\
    Occupation: \texttt{\{occupation\}}\\
Information about  \texttt{\{name2\}}: \\
    Background: \texttt{\{background2\}} \\
    Occupation: \texttt{\{occupation2\}}\\
Both participants are from the \texttt{\{culture\}} culture, you are an expert in \texttt{\{culture\}} culture.\\
\midrule
User Prompt \\
\hdashline
Please predict \texttt{\{name\}}'s intent in the last turn based on the provided conversation, and reason the prediction with respect to the social or cultural expectations in \texttt{\{culture\}} that might influence the tone and content of this interaction in a short sentence. Don't explain if you are unsure of the reasons, only explain if you are very certain, keep it short. \\
Please follow the schema: \\
INTENT: \texttt{\{intent\}}  \\
Please only output the response in English: \\
\bottomrule
\end{tabular}}
\caption{Prompts used to generate intents.}
\label{tab:intent_prompt}
\end{table}

\begin{table}[]
\centering
\resizebox{\columnwidth}{!}{
\begin{tabular}{p{0.95\columnwidth}} 
    \toprule
     Meta Filter (User) Prompt\\
     \midrule
     Please critique the previous judgments and output a meta label on the conversation's alignment with the {culture} culture and the confidence.
     Meta label choices: 1. good, 2. bad \\
     Confidence choices: 1. very confident, 2. confident, 3. not sure \\
     Here is the dialogue: \texttt{\{dialogue\}} \\
     Previous judgements: \texttt{\{judgements\}} \\
     Please output the choice number only (don't explain) using the following schema: \\
     Meta label: <choice> \\
     Confidence: <choice> \\
     Critic: <critic> \\
     \bottomrule
    \end{tabular}}
    \caption{This is the prompt used for judging the quality of the data after the data is evaluated based on the prompt in Table \ref{tab:filter_prompt}.}
    \label{tab:filter_prompt2}
\end{table}

\begin{table}[]
\centering
\resizebox{\columnwidth}{!}{
\begin{tabular}{p{0.95\columnwidth}} 
    \toprule
    Filter (User) Prompt 1\\
    \midrule
     Please read the provided dialogue between two people and their basic information, judge if their conversation aligns with the \texttt{\{culture\}} culture. Output the culture alignment and the confidence.\\
     Culture alignment choices: 1. aligned to the culture, 2. not aligned to the culture \\
     Confidence choices: 1. very confident, 2. confident, 3. not sure \\
     Here is the basic information of the participants in this conversation:  \texttt{\{participants\}} \\
     Here is the dialogue: \texttt{\{dialogue\}} \\
     Please output the choice number only (don't explain) using the following schema:\\
     Culture alignment: <choice> \\
     Confidence: <choice> \\
     \midrule
     Filter (User) Prompt 2\\
     \midrule
     Please read the provided dialogue between two people and their basic information, judge the quality of their conversation. Output quality and confidence. The conversation is bad quality if it contains many repeated sentences toward the end or if the content doesn't align with the given setting.\\
     Quality choices: 1. good, 2. bad \\
     Confidence choices: 1. very confident, 2. confident, 3. not sure \\
     Here is the basic information of the participants in this conversation: \texttt{\{participants\}} \\
     Here is the dialogue: \texttt{\{dialogue\}} \\
     Please output the choice number only (don't explain) using the following schema: \\
     Quality: <choice> \\
     Confidence: <choice> \\
     \bottomrule
    \end{tabular}}
    \caption{Prompts used for evaluating the quality of generated dialogues. The first prompt assesses the cultural alignment of the generated data, the second prompt assesses the general generation quality.}
    \label{tab:filter_prompt}
\end{table}

\end{document}